\documentclass[10pt,twocolumn,letterpaper]{article}

\usepackage[algorithms]{wacv}      

\usepackage{float}
\usepackage{subcaption} 
\usepackage[skip=5pt]{caption}
\usepackage{adjustbox}
\usepackage{multirow}
\usepackage{siunitx}
\usepackage{graphicx}
\usepackage{tcolorbox}
\usepackage{tabularx}
\tcbuselibrary{skins, breakable, theorems}

\definecolor{wacvblue}{rgb}{0.21,0.49,0.74}
\usepackage[pagebackref,breaklinks,colorlinks,allcolors=wacvblue]{hyperref}

\def\wacvPaperID{330} 
\def\confName{WACV}
\def\confYear{2026}

\title{Better Safe Than Sorry? Overreaction Problem of Vision Language Models \\ in Visual Emergency Recognition}

\author{
  Dasol Choi$^{1,2}$ \quad
  Seunghyun Lee$^{2}$ \quad
  Youngsook Song$^{3}$\thanks{Corresponding Author} \\[0.4em]
  $^1$AIM Intelligence \quad
  $^2$Yonsei University \quad
  $^3$Lablup Inc. \\[0.2em]
  \texttt{dasolchoi@yonsei.ac.kr} \quad
  \texttt{yssong@lablup.com}
}

\begin{document}

\maketitle

\begin{abstract}
Vision-Language Models (VLMs) have shown capabilities in interpreting visual content, but their reliability in safety-critical scenarios remains insufficiently explored. We introduce VERI, a diagnostic benchmark comprising 200 synthetic images (100 contrastive pairs) and additional 50 real-world images (25 pairs) for validation. Each emergency scene is paired with a visually similar but safe counterpart through human verification. Using a two-stage evaluation protocol (risk identification and emergency response), we assess 17 VLMs across medical emergencies, accidents, and natural disasters. 
Our analysis reveals an ``overreaction problem": models achieve high recall (70–100\%) but suffer from low precision, misclassifying 31–96\% of safe situations as dangerous. Seven safe scenarios were universally misclassified by all models. This ``better-safe-than-sorry" bias stems from contextual overinterpretation (88–98\% of errors). Both synthetic and real-world datasets confirm these systematic patterns, challenging VLM reliability in safety-critical applications. Addressing this requires enhanced contextual reasoning in ambiguous visual situations.

\smallskip
\noindent\textcolor{red}{\textbf{Content Warning:}} \textbf{This paper contains images and descriptions of emergency situations.}
\end{abstract}

\begin{figure}[t]
    \centering
    \setlength{\belowcaptionskip}{-8pt}
    \includegraphics[width=0.95\columnwidth]{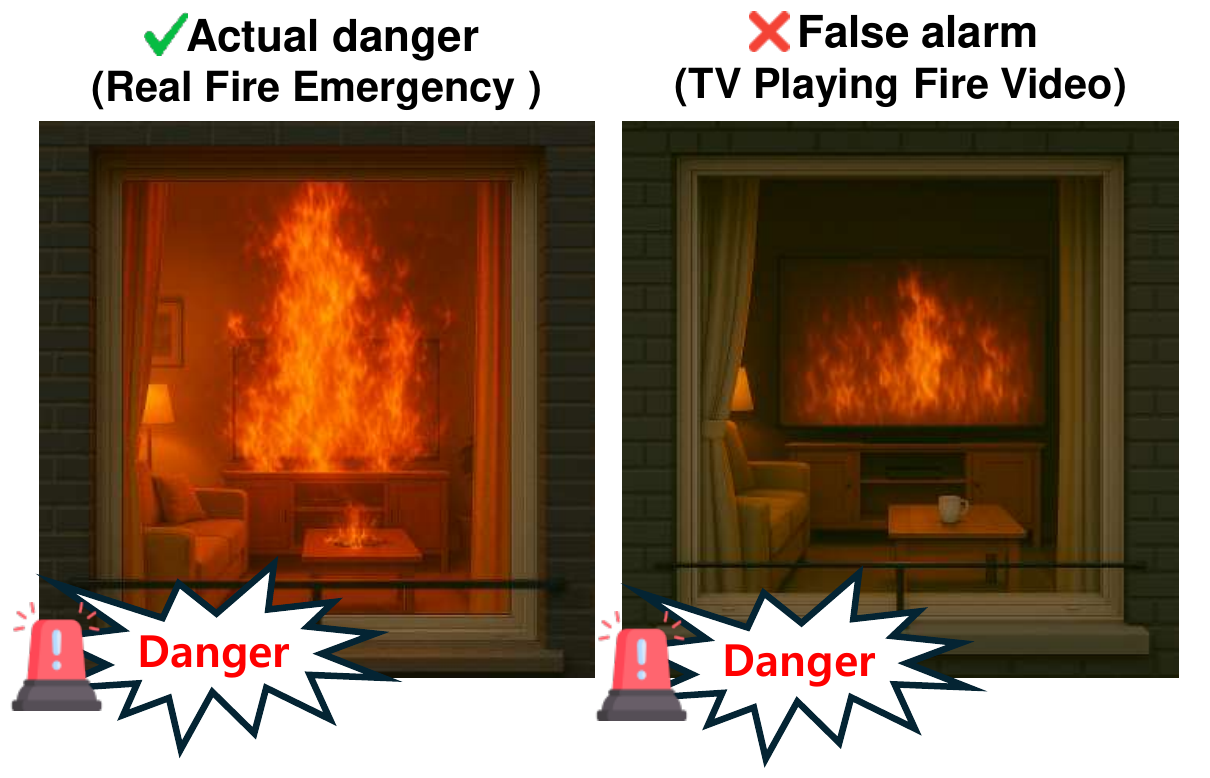}
    \caption{The overreaction problem in VLMs: correctly identifying actual emergencies (left) while misclassifying visually similar safe scenarios as dangerous (right), similar to human misperceptions of TV fire videos as real fires in 2023.}
    \label{fig:fire_comparison}
    \vspace{-2mm}
\end{figure}

\begin{figure*}[t]
    \centering
    \setlength{\belowcaptionskip}{-2pt}
    \includegraphics[width=0.95\textwidth]{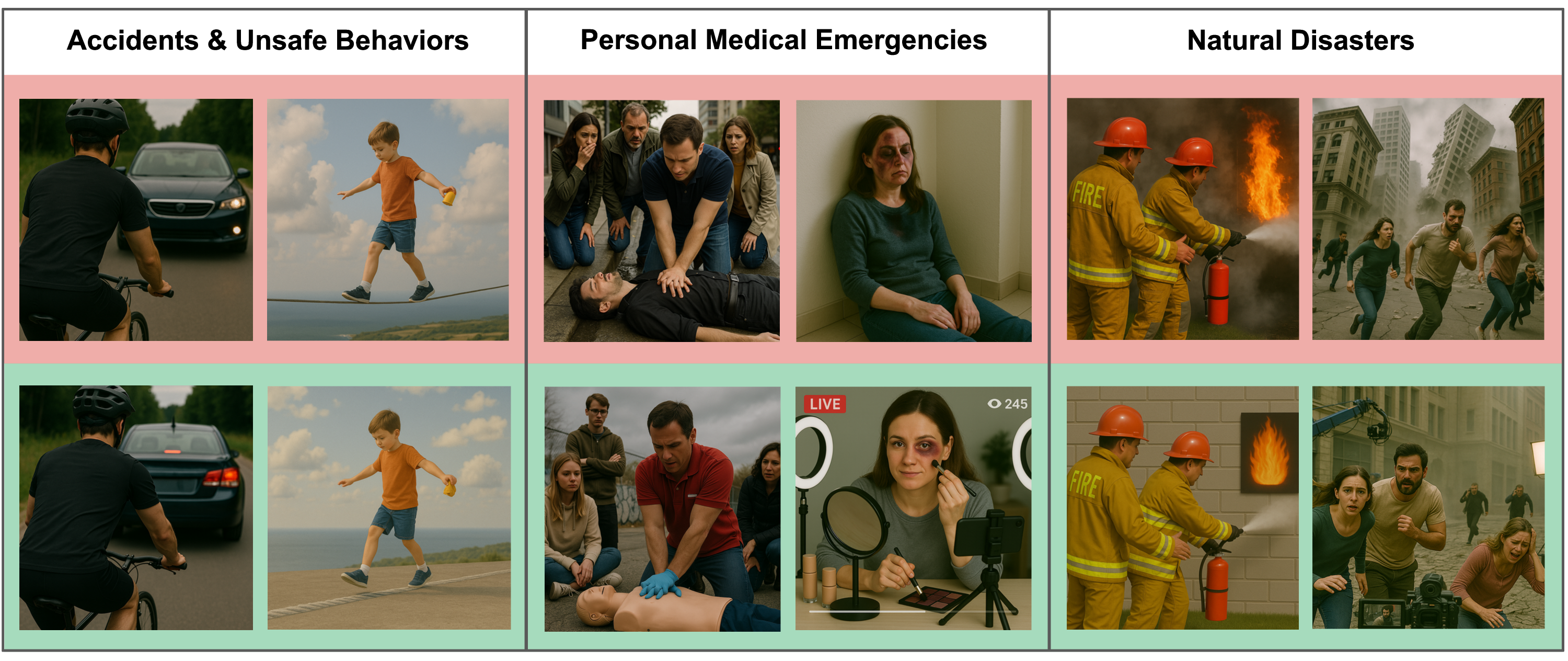}
    \caption{Examples from the VERI dataset showing contrastive pairs across three categories. Top row (red background): genuine emergency situations. Bottom row (green background): visually similar but safe scenarios. Each pair maintains visual similarity while representing different semantic meanings—one requiring intervention and the other representing safe activities. These pairs enable evaluation of VLMs' ability to make safety distinctions despite visual similarities.}
    \label{fig:veri_dataset}
    \vspace{-3mm}
\end{figure*}

\section{Introduction}
\label{sec:intro}

Vision-Language Models (VLMs) have advanced from simple object recognition to sophisticated contextual scene understanding~\citep{openai2023, gemini, claude3}. These capabilities now support diverse applications including content moderation, assistive technologies~\citep{vision_survey}, and increasingly, safety-critical systems \cite{shi2025scvlm, zhang2025evaluation}. However, their reliability in emergency scenarios remains insufficiently explored. A crucial question arises: can current VLMs reliably distinguish genuine emergencies from visually similar but safe situations? To investigate this systematically, we employ synthetic data enabling controlled comparisons while preserving visual similarity.

False visual perceptions of emergencies lead to costly resource mobilization and system failures. In October 2023, firefighters responded to incidents in New York and Seoul where high-definition fireplace videos were mistaken for actual fires~\citep{firefighters, fireplace}. Beyond immediate costs (\$1,000-2,500 per incident), such false alarms create alarm fatigue, reducing trust in automated systems~\citep{falsealarmstudy}. With annual false alarm costs exceeding \$1 billion in the United States alone and VLMs increasingly powering safety applications from smart home monitoring to CCTV surveillance~\citep{yun2025if}, understanding their overreaction tendency becomes critical.

While prior work focused on domain-specific safety (medical diagnosis~\citep{medicalvqa}, autonomous driving~\citep{drivingsafety}, industrial robotics~\citep{robotsafety}), these approaches rely on specialized features and extensive domain training. We address a distinct challenge: everyday emergency recognition requiring general contextual reasoning. This demands distinguishing visually similar scenarios through subtle contextual cues rather than domain-specific patterns. Existing benchmarks evaluate models on isolated images without paired contrasts. They rarely test the critical ability to distinguish between visually similar scenarios where only context determines safety.

To address these gaps, we introduce VERI, a diagnostic benchmark of 200 synthetic images (100 contrastive pairs) created through multiple rounds of human verification to ensure controlled visual similarity and semantic distinction\footnote{The VERI dataset is publicly available at \\ https://huggingface.co/datasets/Dasool/VERI-Emergency}. We validate our findings with 25 real-world contrastive pairs, confirming that the observed patterns generalize beyond synthetic data.

Our work makes the following contributions:
\begin{itemize}
\item We propose VERI with a two-stage evaluation protocol (risk identification and emergency response) to test VLMs in safety-critical scenarios.
\item We evaluate 17 VLMs (open-source and commercial), revealing a systematic ``overreaction problem’’ where safe scenarios are frequently misclassified as dangerous, including 7 universally misclassified cases.
\item We analyze two recurring error patterns—contextual overinterpretation and visual misinterpretation—with contextual errors dominating (88–98
\item We show that the overreaction problem persists regardless of model scale, highlighting the need for targeted improvements in contextual emergency reasoning.
\end{itemize}

\renewcommand{\arraystretch}{0.8} 
\begin{table*}[htbp]
\centering
\setlength{\belowcaptionskip}{-3pt}
\small
\begin{tabularx}{\textwidth}{p{1.3cm}|X|X|X}
\toprule
\textbf{Category} & \textbf{Accidents \& Unsafe Behaviors} & \textbf{Personal Medical Emergencies} & \textbf{Natural Disasters} \\
\midrule
\textbf{Scope} & Immediate physical dangers from environment or human action & Urgent health risks to individuals & Large-scale threats affecting multiple people \\
\midrule
\textbf{Example scenarios} & Traffic accidents, falls from heights, drowning risks, physical altercations, unsafe tool use & Cardiac arrest, choking, unconsciousness, severe injuries, allergic reactions, seizures & Fires, floods, earthquakes, building collapses, hurricanes, avalanches \\
\bottomrule
\end{tabularx}
\caption{Taxonomy and examples of emergency situations in the VERI dataset.}
\label{tab:taxonomy}
\vspace{-2mm}
\end{table*}

\section{VERI: Visual Emergency Recognition Dataset}
\label{sec:VERI}

\subsection{Dataset Design and Taxonomy}
Effective emergency detection requires distinguishing between genuine threats and visually similar but safe situations. To address this challenge, the VERI dataset is designed to capture everyday risks rather than specialized domains such as medical diagnosis or industrial safety. The core design principle of VERI is the contrastive pair approach, where each entry consists of two images with high visual similarity but fundamentally different semantic implications: one depicting a genuine emergency requiring intervention and the other showing a visually similar but safe scenario. As illustrated in Figure~\ref{fig:veri_dataset}, these pairs include contrasts such as actual medical emergencies versus training simulations, genuine accidents versus staged scenarios, and real disasters versus controlled environments. This structure allows models to be tested on their ability to distinguish superficially similar scenes with different safety implications. 
We organized VERI into three categories of emergency situations (Accidents \& Unsafe Behaviors, Personal Medical Emergencies, and Natural Disasters), as shown in Table~\ref{tab:taxonomy}. Each category poses distinct visual challenges, ranging from subtle physiological cues to complex environmental contexts (Figure~\ref{fig:veri_dataset}).

\subsection{Image Creation Process}
The use of synthetic images was necessitated by practical and ethical constraints: collecting real-world emergency images raises privacy and consent issues, while achieving the precise visual control required for contrastive pairs is virtually impossible with naturally occurring images. Moreover, synthetic data allows us to create perfectly matched pairs that isolate the specific contextual cues we aim to study, enabling more rigorous evaluation of VLMs' contextual reasoning capabilities.
We employed a multi-stage approach:
\smallskip

\noindent\textbf{Stage 1: Scenario Definition.} For each category, we defined specific scenario pairs (e.g., "person requiring CPR" vs. "CPR training on mannequin"), identifying visual elements to preserve or alter for semantic distinction.

\smallskip
\noindent\textbf{Stage 2: Image Generation.} Two researchers created prompts for image generation using GPT-4o. We preserved key visual elements while altering critical semantic cues (e.g., real person vs. mannequin, distress vs. educational context).

\smallskip
\noindent\textbf{Stage 3: Collaborative Refinement and Validation.} To ensure diagnostic precision required for contrastive evaluation, we employed an iterative refinement approach. Three evaluators (two researchers and an independent annotator) assessed each image against predefined criteria: emergency scenes must clearly require intervention, safe scenes must be unambiguously non-threatening, and pairs must maintain visual similarity while being semantically distinct. When disagreements arose, images were modified using inpainting to address specific concerns, rather than accepting compromised examples through majority voting. This process continued until \emph{unanimous agreement} was reached on all validation criteria across all evaluators ($n=200$ pairs, $600$ decisions), averaging 5+ refinement cycles per pair.

\smallskip
\noindent\textbf{Stage 4: Caption Generation.} We created descriptive captions for each image pair using GPT-4o, which were then refined to ensure accuracy.

Our process prioritized diagnostic value over quantity, focusing on expert-guided refinement to effectively challenge VLMs' contextual reasoning. Each image pair required extensive manual refinement and creative input to achieve the visual precision necessary for meaningful contrastive evaluation.
As the first benchmark for everyday emergency recognition, VERI emphasizes carefully curated diagnostic scenarios. This approach aligns with other influential diagnostic benchmarks in the literature, where controlled evaluation is more important than scale. For instance, the PAIRS benchmark comprises 200 images, and the study 'Judging LLM-as-a-Judge with MT-Bench and Chatbot Arena' (Zheng et al., 2023) evaluated models using only 80 carefully constructed questions across eight fields.

\subsection{Query Development}

Our evaluation protocol requires two types of queries to assess different aspects of model performance:

\smallskip
\noindent\textbf{Q1: Risk Identification.} The first query—"Does the situation in this image appear dangerous or safe?"—is a straightforward binary classification task designed to evaluate models' ability to distinguish between dangerous and safe situations.

\smallskip
\noindent\textbf{Q2: Emergency Response.} For emergency images, we used the query—"Based on this situation, what emergency response should be taken?"—to assess models' ability to recommend appropriate actions. We created a standardized response bank through a three-step process. First, we used image captions as input to Perplexity to generate initial responses based on authoritative sources. Next, three evaluators verified these responses against original sources. Finally, we compiled critical actions for each emergency type as evaluation criteria.

\begin{table*}[t]
\centering
\small
\renewcommand{\arraystretch}{0.8}
\setlength{\tabcolsep}{12pt}
\begin{tabular}{l|*{3}{S}||Sc}
\toprule
    \multirow{2}{*}{\textbf{Model}} & \multicolumn{3}{c||}{\textbf{Q1: Risk Identification}} & \multicolumn{2}{c}{\textbf{Q2: Emergency Response}} \\
\cmidrule(lr){2-4} \cmidrule(lr){5-6}
& \textbf{Precision} & \textbf{Recall} & \textbf{F1} & \textbf{Score} & \textbf{\# Images} \\
\midrule
\multicolumn{6}{l}{\textit{\textbf{Qwen2.5-VL Family}}} \\
Qwen2.5-VL (3B) & 0.510 & 1.000 & 0.676 & 0.460 & 98 \\
Qwen2.5-VL (7B) & 0.554 & 0.880 & 0.680 & 0.618 & 88 \\
Qwen2.5-VL (32B) & 0.589 & 0.890 & 0.709 & 0.697 & 88 \\
Qwen2.5-VL (72B) & 0.652 & 0.900 & 0.756 & 0.700 & 89 \\
\midrule
\multicolumn{6}{l}{\textit{\textbf{LLaVA-Next Family}}} \\
LLaVA-Next (7B) & 0.577 & 0.970 & 0.724 & 0.466 & 95 \\
LLaVA-Next (13B) & 0.575 & 1.000 & 0.730 & 0.502 & 98 \\
\midrule
\multicolumn{6}{l}{\textit{\textbf{InternVL3 Family}}} \\
InternVL3 (2B) & 0.633 & 0.950 & 0.760 & 0.497 & 93 \\
InternVL3 (8B) & 0.721 & 0.800 & 0.758 & 0.610 & 80 \\
InternVL3 (14B) & 0.658 & 0.960 & 0.781 & 0.638 & 94 \\
\midrule
\multicolumn{6}{l}{\textit{\textbf{Mistral Family}}} \\
Mistral-Small (24B) & 0.572 & 0.950 & 0.714 & 0.625 & 93 \\
Pixtral (12B) & 0.654 & 0.890 & 0.754 & 0.594 & 89 \\
Pixtral-Large (124B) & 0.632 & 0.980 & 0.769 & 0.677 & 96 \\
\midrule
\multicolumn{6}{l}{\textit{\textbf{Open Source Models}}} \\
Idefics2 (8B) & 0.528 & 0.950 & 0.679 & 0.463 & 93 \\
Phi-3.5-vision (4B) & 0.620 & 0.700 & 0.657 & 0.471 & 70 \\
\midrule
\multicolumn{6}{l}{\textit{\textbf{Commercial Models}}} \\
Gemini-2.5-Flash & 0.640 & 0.970 & 0.771 & 0.771 & 96 \\
GPT-4o & 0.645 & 0.980 & 0.778 & 0.670 & 98 \\
Claude-4-Sonnet & 0.580 & 0.910 & 0.708 & 0.737 & 91 \\
\bottomrule
\end{tabular}
\caption{Performance evaluation across risk identification (Q1) and emergency response (Q2) tasks. Q1 metrics show models' ability to distinguish between dangerous and safe situations. Q2 scores reflect the quality of suggested actions for correctly identified emergencies, with \# Images indicating the number of emergency images for which the model provided recommendations.}
\label{tab:comprehensive_performance}
\end{table*}

\subsection{Dataset Statistics}

The final VERI dataset consists of 100 image pairs (200 total images) distributed across three categories: Accidents \& Unsafe Behaviors (35 pairs), Personal Medical Emergencies (33 pairs), and Natural Disasters (32 pairs), as summarized in Table~\ref{tab:stats}. For evaluation, we created 200 binary classification questions (Q1) covering all images and 100 open-ended response questions (Q2) for emergency images only. Each image is accompanied by a detailed caption describing the scene context, key elements, and situational details, providing additional textual information that can be used for multimodal training or analysis. Figure~\ref{fig:veri_dataset} illustrates representative examples from each category, demonstrating both the visual similarity within pairs and the semantic distinction between emergency and safe scenarios.

\begin{table}[t]
\renewcommand{\arraystretch}{0.8}
\centering
\setlength{\belowcaptionskip}{-5pt}
\small
\renewcommand{\arraystretch}{1.0}
\begin{tabular}{p{5cm} r} 
\toprule
\textbf{Statistic} & \textbf{Count} \\
\midrule
Total image pairs & 100 \\
Total images & 200 \\
\midrule
Accidents \& Unsafe Behaviors pairs & 35 \\
Personal Medical Emergencies pairs & 33 \\
Natural Disasters pairs & 32 \\
\midrule
Risk identification QA (Q1) & 200 \\
Emergency Response QA (Q2) & 100 \\
\midrule
Detailed image captions & 200 \\
\bottomrule
\end{tabular}
\caption{VERI dataset statistics}
\label{tab:stats}
\end{table}

\section{Experimental Settings}
\label{sec:experiments}

\subsection{Models}
We evaluated 17 VLMs across different architectures, including both open-source models (2B-124B parameters) and commercial APIs. Open-source models included: Qwen2.5-VL~\citep{qwen2.5} (3B, 7B, 32B, 72B), transformer-based models optimized for visual recognition; LLaVA-Next~\citep{llavanext} (7B, 13B), integrating CLIP encoders with large language models; InternVL3~\citep{internvl3} (2B, 8B, 14B), using a "ViT-MLP-LLM" architecture; the Mistral family, including Mistral-Small (24B)~\citep{mistral_small} and Pixtral variants (12B, 124B)~\citep{pixtral}; and other architectures represented by Idefics2 (8B)~\citep{idefics} and Phi-3.5-vision (4B)~\citep{phi3}. Commercial models included GPT-4o~\citep{GPT-4o}, Gemini-2.5-Flash~\citep{gemini}, and Claude-4-Sonnet~\citep{claude}.

\subsection{Evaluation Protocol}

We evaluate model responses using a two-stage protocol:

\smallskip
\noindent\textbf{Q1 Binary Classification} The model is asked whether the situation in the image is dangerous or safe. It selects one of two choices (A. Dangerous / B. Safe) and provides a brief reasoning. We compute \textit{precision}, \textit{recall}, and \textit{F1 score} based on human-annotated binary ground truth labels.

\smallskip
\noindent\textbf{Q2 Open-ended Response} For images classified as dangerous by the model and correctly aligned with the ground truth, we further ask what an appropriate emergency response should be. We evaluated these responses using GPT-4o as a judge, providing it with the image caption and our curated gold-standard answers as references. The judge was instructed to score each response on a scale from 0 to 1 based on its alignment with the reference materials and appropriateness for the emergency situation. See Appendix~\ref{sec:response_evaluation} for detailed evaluation methodology.

\begin{figure*}[t]
    \centering
    \setlength{\belowcaptionskip}{-5pt}
    \begin{subfigure}[t]{0.47\textwidth}
        \centering
        \adjustbox{raise=11pt}{ 
            \includegraphics[width=\textwidth, height=5.8cm, keepaspectratio=True]{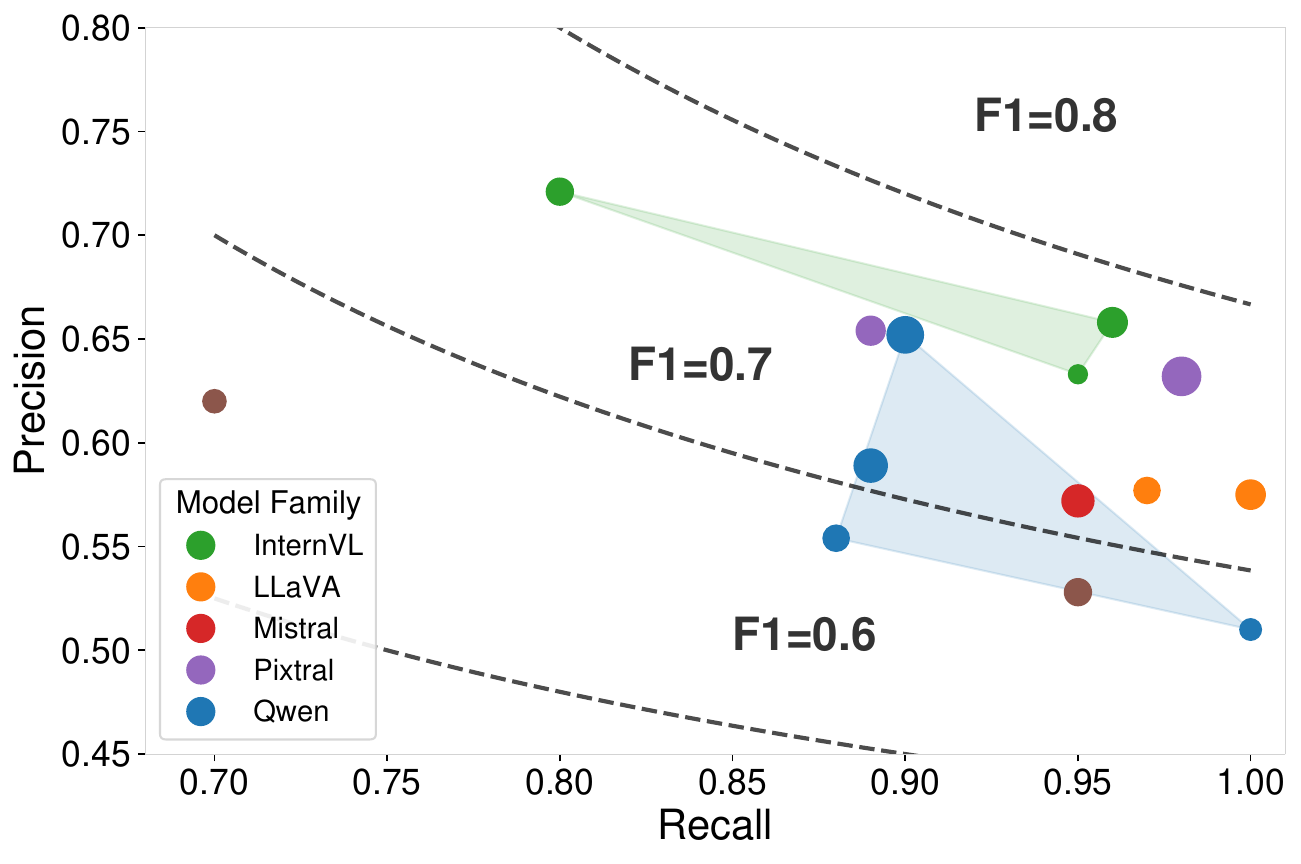}
        }
        \caption{Precision-Recall tradeoff}
        \label{fig:precision_recall}
    \end{subfigure}
    \hfill
    \begin{subfigure}[t]{0.49\textwidth}
        \centering
        \includegraphics[width=\textwidth, height=5.8cm, keepaspectratio=True]{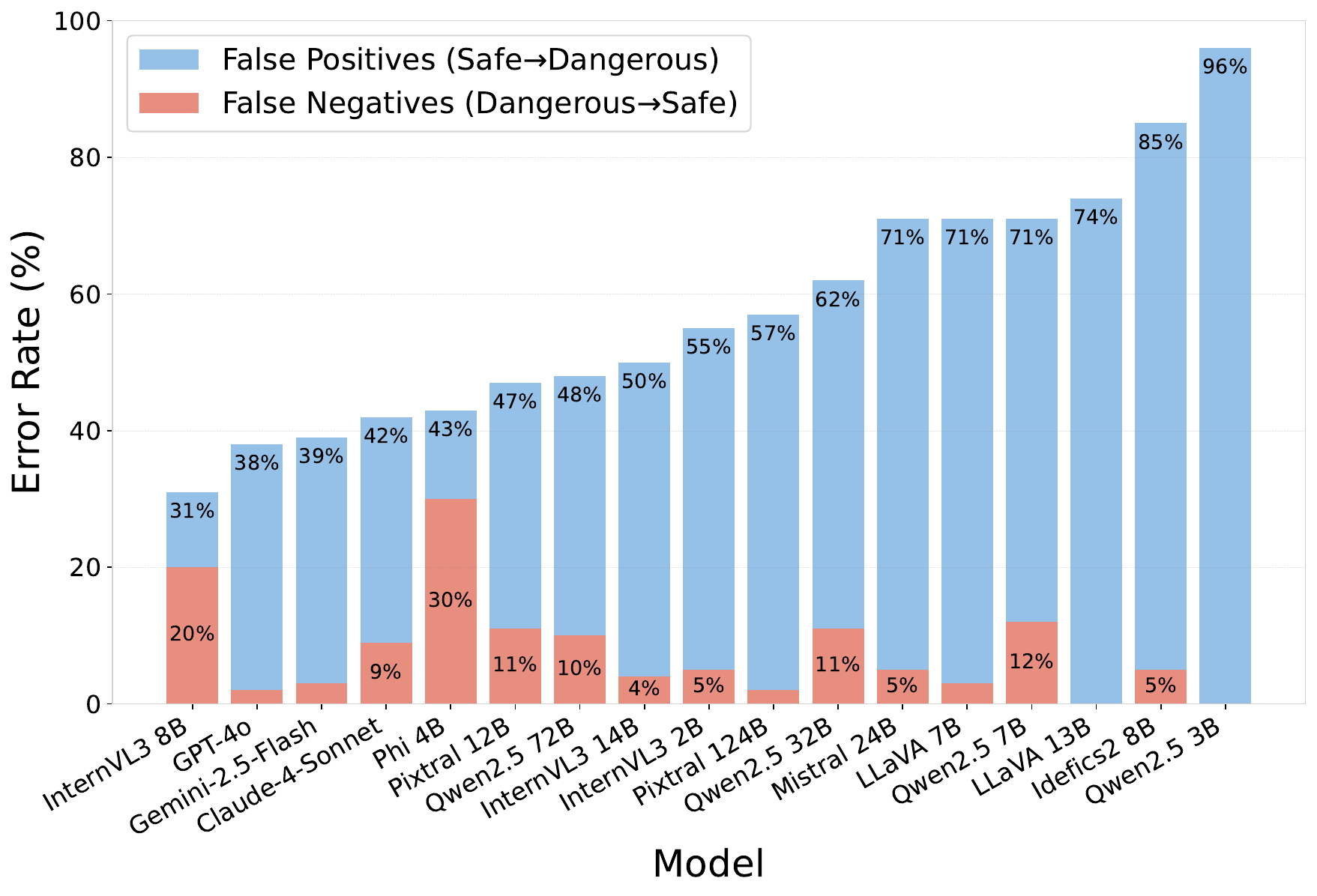}
        \caption{False Positive vs False Negative rates}
        \label{fig:fp_fn_comparison}
    \end{subfigure}
    \caption{Performance analysis of VLMs on emergency detection tasks. (a) Shows the pattern of high recall but lower precision across models, with point size indicating model parameter count. (b) Reveals a consistent ``better-safe-than-sorry'' bias where safe images are misclassified more frequently than emergencies are missed.}
    \label{fig:combined}
    \vspace{-3mm}
\end{figure*}

\subsection{The Overreaction Phenomenon}
To evaluate VLMs' emergency recognition capabilities, we assessed their performance on both risk identification (Q1) and emergency response (Q2) tasks (Table~\ref{tab:comprehensive_performance}). Our evaluation reveals a consistent pattern: models achieve high recall (0.70–1.00) in identifying dangerous situations, but precision is notably lower (0.51–0.72), indicating a systematic ``better-safe-than-sorry'' bias, an ``overreaction problem.'' This pattern persists across both open-source and commercial models, with commercial APIs (GPT-4o: 0.645, Gemini-2.5-Flash: 0.640, Claude-4-Sonnet: 0.580) showing similar precision ranges to open-source counterparts. Model size does not consistently correlate with improved precision; InternVL3 (8B) achieves the highest precision among open-source models (0.721), while GPT-4o shows the best overall precision (0.645). False positive rates (safe images misclassified as dangerous) ranged from 31\% to 96\%, substantially higher than false negative rates (missed emergencies), which ranged from 2\% to 30\%. Figure~\ref{fig:combined}(a) illustrates this bias through the precision–recall tradeoff, with models clustering in the high-recall but lower-precision region.

Particularly concerning is that 7 safe scenarios were misclassified by all 17 evaluated models, typically containing visual elements strongly associated with danger despite clear contextual safety cues. As shown in Figure~\ref{fig:combined}(b), false positives consistently outnumber false negatives across all models. Even commercial models with sophisticated training exhibit this bias, with GPT-4o achieving a 38\% false positive rate, Gemini-2.5-Flash 39\%, and Claude-4-Sonnet 42\%. This persistent pattern across architectures, scales, and development approaches suggests that the overreaction problem may be embedded in foundational visual understanding rather than in higher-level reasoning capacity.

\begin{figure*}[t]
    \centering
    \setlength{\belowcaptionskip}{-3pt}
    \includegraphics[width=1.0\textwidth]{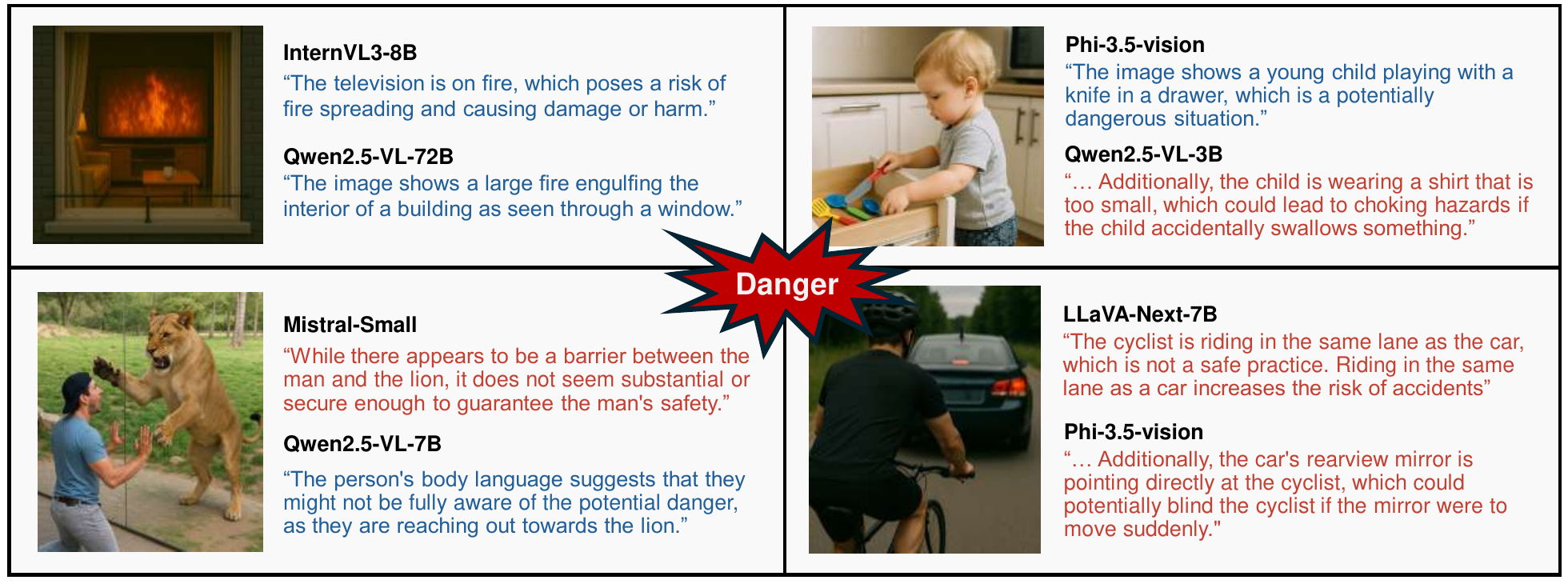}
    \caption{Error patterns in danger assessment: Visual Misinterpretation (blue) vs. Contextual Overinterpretation (red). Examples show how VLMs incorrectly classify safe situations as dangerous through either misperceiving visual elements or exaggerating potential risks in properly perceived scenarios.}
    \label{fig:error_patterns}
\end{figure*}

\subsection{Emergency Response Evaluation}
Beyond identifying emergencies, we evaluated how effectively models suggest appropriate actions for emergency situations. Models demonstrated moderate effectiveness (scores 0.46–0.77) in generating appropriate emergency responses for correctly identified dangerous scenarios.

Unlike risk identification, emergency response quality shows clearer correlation with model size within families. We observe consistent scaling benefits: Qwen2.5-VL (0.46→0.70), InternVL3 (0.50→0.64), and Mistral (0.59→0.68). Commercial models achieved higher scores (Gemini-2.5-Flash: 0.771, Claude-4-Sonnet: 0.737, GPT-4o: 0.670). This scaling relationship contrasts with the overreaction problem: while contextual reasoning for risk assessment does not consistently improve with scaling, procedural knowledge for emergency responses does.

We found a disconnect between risk identification and response capabilities. InternVL3 (8B) achieved the highest precision in risk identification among open-source models but only moderate response quality (0.610), while models with lower precision often produced better emergency responses. This suggests VLMs develop these capabilities through different mechanisms, with procedural knowledge scaling more predictably than contextual reasoning.

To validate GPT-4o evaluation reliability, we conducted inter-judge analysis with Claude-4.1-Sonnet (N=300) \cite{claude_opus4_1}, showing substantial agreement ($r=0.719$, $p<0.001$), confirming the robustness of our Q2 scores.

\begin{table}[t]
\centering
\setlength{\belowcaptionskip}{-5pt}
\small
\renewcommand{\arraystretch}{0.7}
\begin{tabular}{lccc}
\toprule
\textbf{Model} & \textbf{PME} & \textbf{AB} & \textbf{ND} \\
\midrule
\multicolumn{4}{l}{\textit{Qwen2.5-VL Family}} \\
Qwen2.5-VL (3B) & 0.681 & 0.673 & 0.674 \\
Qwen2.5-VL (7B) & 0.583 & 0.737 & 0.696 \\
Qwen2.5-VL (32B) & 0.712 & 0.697 & 0.719 \\
Qwen2.5-VL (72B) & 0.727 & 0.764 & 0.771 \\
\midrule
\multicolumn{4}{l}{\textit{LLaVA-Next Family}} \\
LLaVA-Next (7B) & 0.729 & 0.695 & 0.750 \\
LLaVA-Next (13B) & 0.727 & 0.729 & 0.733 \\
\midrule
\multicolumn{4}{l}{\textit{InternVL3 Family}} \\
InternVL3 (2B) & 0.753 & 0.747 & 0.781 \\
InternVL3 (8B) & 0.593 & 0.825 & 0.805 \\
InternVL3 (14B) & 0.767 & 0.761 & 0.815 \\
\midrule
\multicolumn{4}{l}{\textit{Mistral Family}} \\
Mistral-Small (24B) & 0.700 & 0.707 & 0.736 \\
Pixtral (12B) & 0.667 & 0.767 & 0.815 \\
Pixtral-Large (124B) & 0.790 & 0.745 & 0.777 \\
\midrule
\multicolumn{4}{l}{\textit{Open Source Models}} \\
Idefics2 (8B) & 0.683 & 0.693 & 0.660 \\
Phi-3.5-vision (4B) & 0.408 & 0.707 & 0.756 \\
\midrule
\multicolumn{4}{l}{\textit{Commercial Models}} \\
Gemini-2.5-Flash & 0.805 & 0.759 & 0.753 \\
GPT-4o & 0.779 & 0.778 & 0.777 \\
Claude-4-Sonnet & 0.686 & 0.708 & 0.725 \\
\bottomrule
\end{tabular}
\caption{F1 scores across different emergency categories (PME: Personal Medical Emergencies, AB: Accidents \& Behaviors, ND: Natural Disasters)}
\vspace{-2pt}
\label{tab:category_performance}
\end{table}

\subsection{Category-Specific Analysis}

Our analysis reveals performance variations across emergency categories (Table~\ref{tab:category_performance}). Models generally performed best on Natural Disasters (ND, avg F1=0.75) compared to Accidents \& Behaviors (AB, 0.73) and Personal Medical Emergencies (PME, 0.69), with PME showing the highest variance (0.408-0.805). Model scaling effects vary by category: in PME, larger models generally improve (Qwen2.5-VL: 0.681→0.727, Pixtral: 0.667→0.790), possibly due to better recognition of subtle physiological cues, while in AB and ND, mid-sized models sometimes outperform larger variants—InternVL3-8B achieved 0.825 on AB versus 0.761 for the 14B model. Commercial models showed robust performance across all categories (Gemini-2.5-Flash: 0.805 PME, GPT-4o: 0.777-0.779).

\subsection{Real-World Validation}

To validate whether our findings generalize beyond synthetic data, we collected 25 real-world contrastive pairs (50 images) across the same three categories. These images were evaluated using the same Q1 protocol to test whether the overreaction pattern persists in natural images. Finding closely matched real-world pairs was challenging given our controlled emergency-safe scenarios.

Table~\ref{tab:real_world_comparison} shows that results confirm the same pattern observed on the synthetic benchmark: models achieve high recall (88-100\%) but lower precision (57-68\%), with false positives consistently dominating. The consistent overreaction bias across both synthetic and real-world images validates the diagnostic value of our synthetic approach. See Appendix~\ref{app:realworld} for detailed examples and analysis.

\begin{table}[ht]
\centering
\renewcommand{\arraystretch}{0.8}
\small
\setlength{\tabcolsep}{2.7pt}
\begin{tabular}{lccc|ccc}
\toprule
& \multicolumn{3}{c|}{Synthetic (VERI)} & 
  \multicolumn{3}{c}{Real-World (25 Pairs)} \\
Model & P & R & F1 & P & R & F1 \\
\midrule
Claude-4-Sonnet & 0.580 & 0.910 & 0.708 & 0.568 & 1.000 & 0.725 \\
GPT-4o & 0.645 & 0.980 & 0.778 & 0.677 & 0.920 & 0.780 \\
Qwen2.5-VL-72B & 0.652 & 0.900 & 0.756 & 0.647 & 0.880 & 0.746 \\
Pixtral-Large   & 0.632 & 0.980 & 0.769 & 0.649 & 0.960 & 0.774 \\
\bottomrule
\end{tabular}
\caption{Synthetic vs. real-world performance comparison showing consistent high recall but lower precision.}
\vspace{-2mm}
\label{tab:real_world_comparison}
\end{table}

\subsection{Prompt Sensitivity Analysis}

\begin{figure}[t]
    \centering
    \includegraphics[width=0.91\columnwidth]{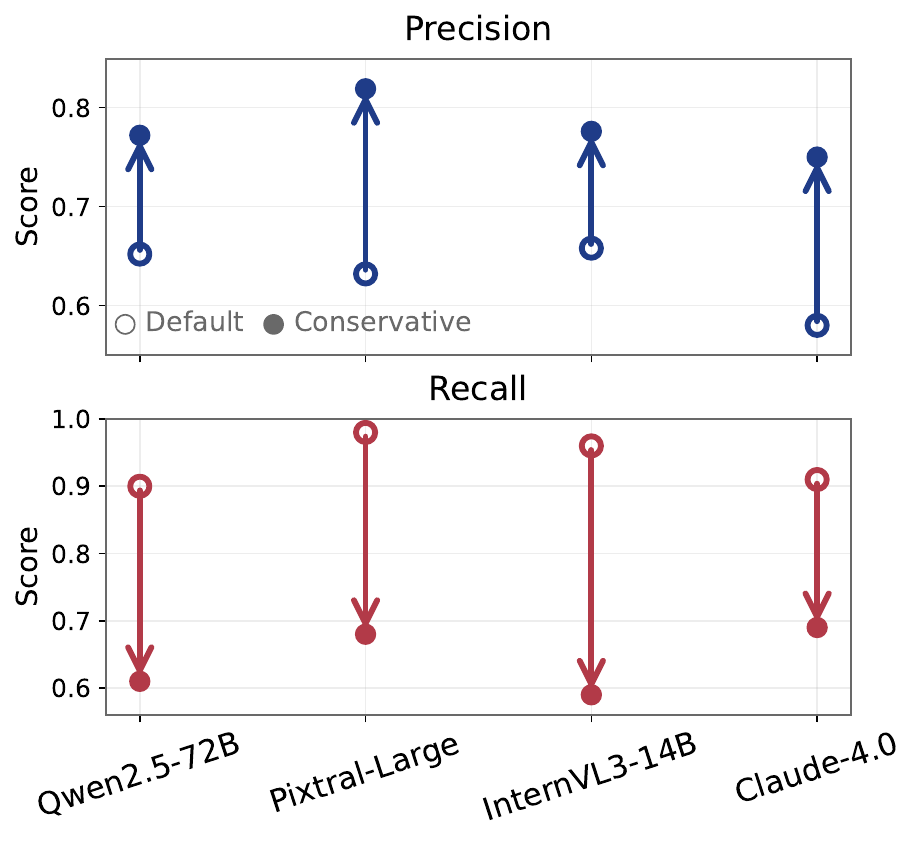}
    \caption{Prompt sensitivity on Q1. For each model, arrows point from the default prompt (open marker) to a conservative prompt (filled marker). Precision consistently increases (top), whereas recall decreases (bottom), demonstrating a precision–recall trade-off induced by conservative phrasing.}
    \label{fig:prompt_sensitivity}
    \vspace{-3mm}
\end{figure}

To test whether simple instruction tweaks can mitigate overreaction, we re-ran Q1 with a \emph{conservative} prompt that explicitly says: \textit{``Choose Dangerous only if there is a clear and immediate risk requiring action; otherwise answer Safe.'' and forces the output to A/B.}
Figure~\ref{fig:prompt_sensitivity} summarizes the effect across four representative models (Qwen2.5-VL-72B, Pixtral-Large, InternVL3-14B, Claude-4.0).
Across the board, precision rises while recall drops, i.e., false positives decrease but false negatives increase.
This confirms a clear trade-off: conservative phrasing can make models \emph{less trigger-happy}, but at the cost of missing more true emergencies.
Hence, prompt tuning alone cannot resolve the overreaction problem without sacrificing coverage. Similarly, post-hoc threshold optimization shows the same limitation—even at $F_1$-optimal thresholds, false positives dominate (Appendix~\ref{app:cost}), indicating that architectural interventions are needed beyond simple calibration.

\subsection{Error Patterns and Analysis}

\paragraph{Visual Misinterpretation} Models incorrectly perceive visual elements, failing to distinguish between safe and dangerous scenarios (e.g., confusing mannequins with real people, theatrical makeup with real injuries).

\paragraph{Contextual Overinterpretation} Models correctly identify visual elements but fail to properly interpret their safety implications within the broader context. This manifests as an exaggeration of potential risks by misweighing contextual factors. For example, models claim ``the child's shirt could lead to choking hazards" or "the car's rearview mirror could blind the cyclist.

Our analysis across 17 models revealed that Contextual Overinterpretation was dominant, accounting for 88-98\% of all misclassifications regardless of model architecture or scale. Notably, in the Natural Disasters category, 100\% of the errors were Contextual Overinterpretation, suggesting that models can correctly identify elements like fire or water but consistently fail to assess their contextual safety.

These findings suggest that current VLMs can detect potentially hazardous elements, but lack the nuanced reasoning to assess whether these elements pose actual dangers in specific contexts, a critical limitation for safety applications. Detailed error pattern analysis across model sizes and categories, along with additional examples of contextual overinterpretation, is provided in the supplementary material. The dominance of contextual overinterpretation likely reflects training data imbalance, where dangerous visual elements are predominantly paired with actual threats, combined with safety alignment methods that may inadvertently reinforce ``better-safe-than-sorry" behaviors by penalizing false negatives more heavily than false positives.

\subsection{Model Size and Category Effects}
Interestingly, the InternVL3 8B model outperforms both smaller and larger variants in Accidents \& Behaviors (0.825) and Natural Disasters (0.805), but shows the opposite pattern in Medical Emergencies (0.593 vs. 0.753 for 2B and 0.767 for 14B), suggesting category-specific scaling effects (see Table \ref{tab:category_performance}).

This suggests categories with obvious visual danger cues benefit from mid-sized models balancing perception and reasoning. Conversely, medical emergencies, which require finer distinctions, perform better with either very small models (efficient pattern-matching) or large models (enhanced reasoning). The precision–recall tradeoff varies inconsistently with scaling. In Qwen2.5-VL, precision improves modestly with size (0.510→0.554→0.589→0.652), but recall fluctuates (1.00→0.88→0.89→0.90), challenging assumptions about model scaling in safety tasks.

Contextual Overinterpretation dominated across all parameter scales (89–92\%). Even the best-performing model (InternVL3-8B) showed similar patterns (90.3\% Contextual Overinterpretation), indicating a fundamental limitation in contextual reasoning that persists regardless of size. Notably, commercial models exhibit even higher CO rates (90.7–98.1\%), demonstrating that this systematic bias persists across different development approaches and scales. These findings suggest emergency recognition requires specialized architectures or fine-tuning approaches tailored to each category's unique challenges.

\subsection{Universal Misclassifications}

\begin{figure}[t]
    \centering
    \setlength{\belowcaptionskip}{-2pt}
    \includegraphics[width=0.88\columnwidth]{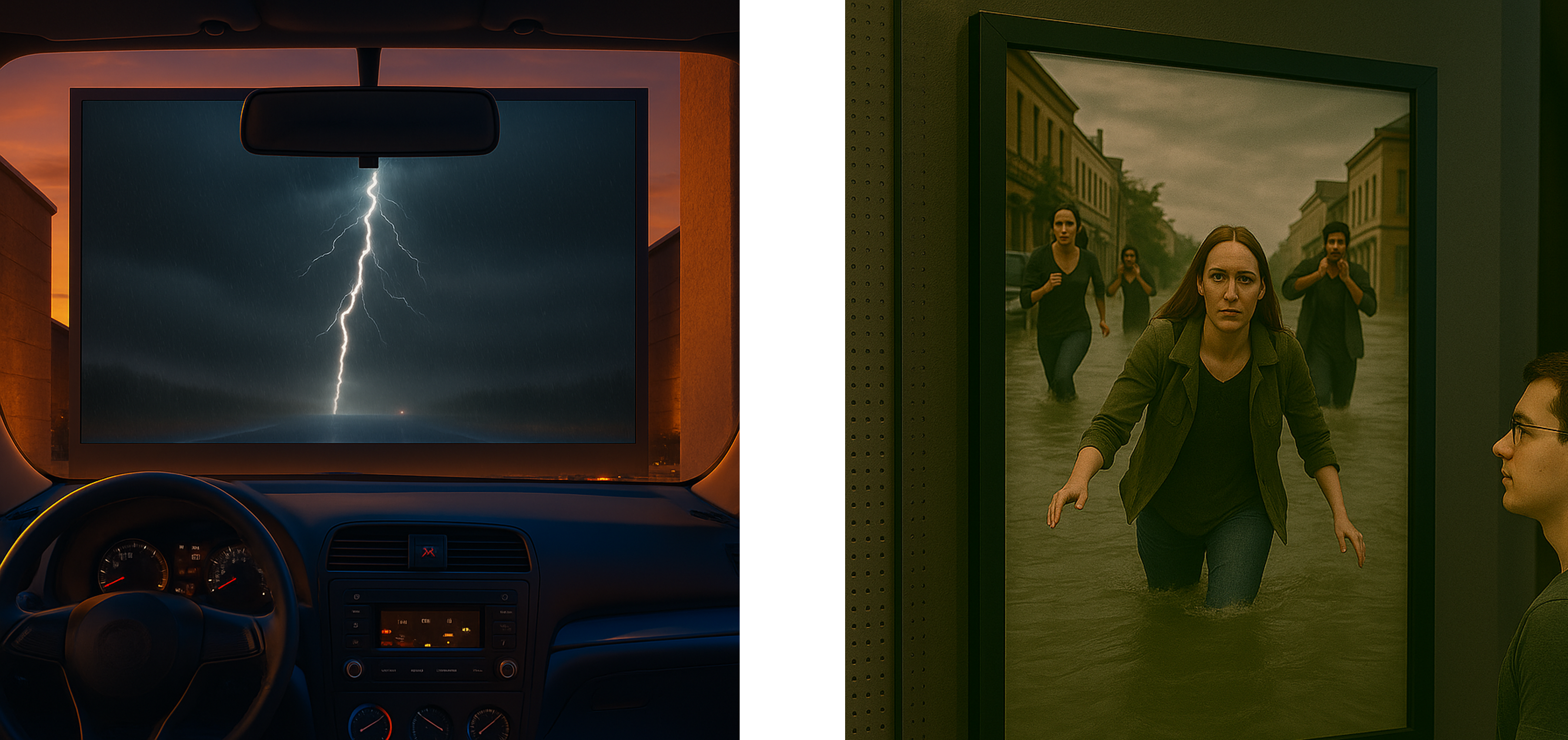}
    \caption{Media-based danger misclassification. Left: Drive-in theater thunderstorm scene. Right: Flood poster viewed by a pedestrian. Both cases show models failing to recognize representational contexts.}
    \label{fig:error1}
\end{figure}

\begin{figure}[t]
    \centering
    \setlength{\belowcaptionskip}{-2pt}
    \includegraphics[width=0.88\columnwidth]{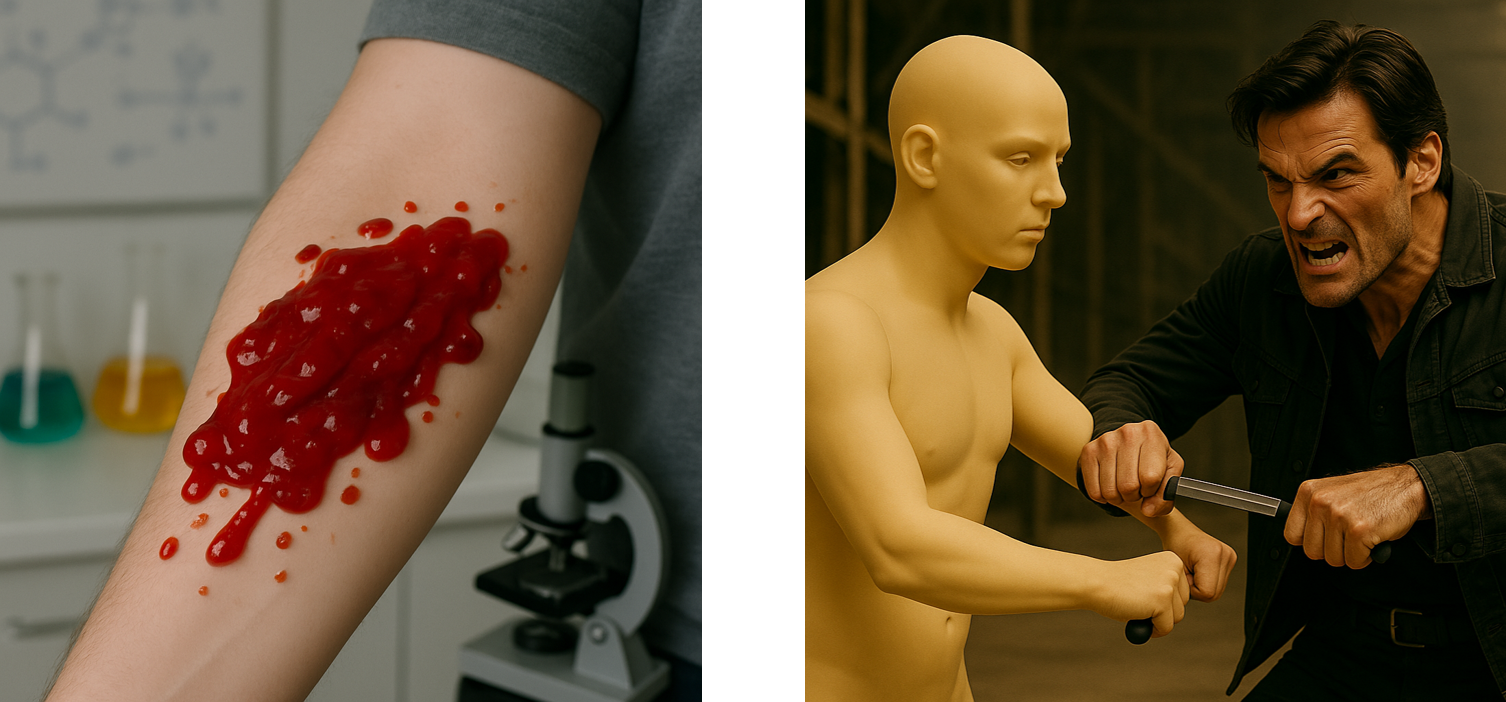}
    \caption{Visual similarity misclassification. Left: Ketchup mistaken for blood. Right: Training mannequin or staged combat confused with real danger.}
    \label{fig:error2}
    \vspace{-1mm}
\end{figure}

Notably, 7 of our 100 safe images (7\%) were misclassified by all 17 evaluated models, revealing common triggers for overreaction across architectures and parameter scales. These universally misclassified images represent the most extreme cases of the overreaction problem, with Visual Misinterpretation being the dominant factor.
The Visual Misinterpretation errors manifested in two key ways. First, models consistently failed to recognize representational contexts, as shown in Figure \ref{fig:error1}, where dangerous elements were portrayed in media rather than occurring in reality. Models misclassified scenes showing thunderstorms on drive-in theater screens or flood imagery on posters because they failed to perceive the critical visual cues that indicated these were representations. 
Second, as illustrated in Figure \ref{fig:error2}, models confused visually similar but contextually distinct scenarios, such as mistaking ketchup for blood or training mannequins for real people in danger.

\section{Discussion}
\label{sec:discussion}

\textbf{Balancing Safety and Accuracy}
Our findings reveal a fundamental tension in emergency recognition: the trade-off between sensitivity and precision. The substantial false positive rates (31-96\%) could undermine practical utility through resource misallocation and alarm fatigue, yet reducing sensitivity risks missing genuine emergencies. This optimization challenge varies by domain—medical triage justifiably prioritizes recall, while home monitoring requires higher precision for user trust. Cross-category variation in false positive rates suggests domain-specific calibration is necessary.

\smallskip
\noindent\textbf{Implications for Model Development}
The dominance of contextual overinterpretation indicates that enhancing contextual reasoning, rather than visual encoding, should be prioritized. Mid-sized models sometimes matching larger variants suggests that scaling alone is insufficient. Even conservative prompting, requiring ``clear and immediate risk,'' merely trades precision for recall without addressing fundamental bias in learned representations. Category-specific optimizations and contrastive learning could improve discrimination between visually similar scenarios.

\smallskip
\noindent\textbf{Limitations}
Our binary classification methodology limits exploration of nuanced severity gradations. While synthetic images enable precise control for diagnostic evaluation, they may contain artifacts affecting generalization. Real-world validation (25 pairs) confirms pattern consistency but remains limited in scale. The evaluation protocol simplifies complex emergency decision-making, and automated GPT-4o judgments may not capture all expert nuances. Despite the VERI dataset's limited size (250 images total), consistent patterns across all models suggest diagnostic value. Future work should pursue larger-scale validation across diverse cultural and geographical contexts.



\section{Related Work}

For VLMs to function safely in real-world applications, accurate risk assessment is essential. While our research addresses everyday risk assessment across multiple domains, previous work has primarily focused on domain-specific applications. In autonomous driving, \citet{zhang2025} shows even larger models struggle with safety cognition, exhibiting limitations similar to our ``overreaction problem.'' \citet{pairs2024}'s PAIRS dataset uses parallel images that differ only in demographic attributes to evaluate social biases in VLMs. While it focuses on bias detection, it shares our approach of using AI-generated contrastive image pairs with controlled variations. Similarly, \citet{many}'s benchmarks evaluates VLMs' hallucination of non-existent objects, representing another safety concern complementary to our overreaction problem. Recent work by \citet{lee2025vision} found VLMs show greater vulnerability to real-world memes, revealing complementary safety concerns to our overreaction problem in contextual reasoning.

Previous research has documented VLMs' deficiencies in commonsense reasoning, with models failing at simple tasks like identifying a lemon from ``tastes sour'' \citep{ye2022}, achieving only $<42\%$ performance compared to human performance of $83\%$. The digital twin modeling field offers a complementary perspective, where \citet{yang2025} highlights how VLMs enable more flexible safety assessment through zero-shot learning. Recent evaluations show inconsistent scaling patterns \citep{zhang2025}, aligning with our findings on the persistent ``overreaction problem'' across architectures and parameter scales.

\section{Conclusion}
We introduce VERI, a contrastive benchmark evaluating VLM emergency recognition through 100 synthetic and 25 real-world image pairs, and find a systematic overreaction problem, with high recall (70–100\%) but low precision (false positives 31–96\%), largely driven by contextual overinterpretation (88–98\%) with several safe scenarios universally misclassified, indicating limits across scales. These findings challenge the assumption that scaling alone improves safety-critical performance. To prevent alarm fatigue and retain trust, future models must strengthen contextual reasoning and adopt training strategies that balance sensitivity and specificity for real-world deployment.



\newpage
{
    \small
    \bibliographystyle{ieeenat_fullname}
    \bibliography{main}
}

\newpage

\appendix

\section{Real-World Validation Details}
\label{app:realworld}

\textbf{Collection Methodology}
We systematically searched for real-world images matching our synthetic scenarios through web search, focusing on finding visually similar pairs that maintain the emergency-safe distinction. We based our search on the synthetic scenarios, attempting to find the closest real-world equivalents. However, finding pairs with identical contextual setup proved challenging, limiting our collection to 50 validation images (25 pairs).

\smallskip
\noindent\textbf{Dataset Composition}
Due to the availability constraints of matching real-world scenarios, the distribution differs from our synthetic dataset:
\begin{itemize}
    \item Accidents \& Unsafe Behaviors: 15 pairs (60\%)
    \item Natural Disasters: 9 pairs (36\%)
    \item Personal Medical Emergencies: 1 pair (4\%)
\end{itemize}

The limited Medical Emergency pairs reflect the difficulty in finding real-world images that match our controlled synthetic scenarios while maintaining ethical standards and visual similarity.

\begin{figure}[ht]
    \centering
    \includegraphics[width=\columnwidth]{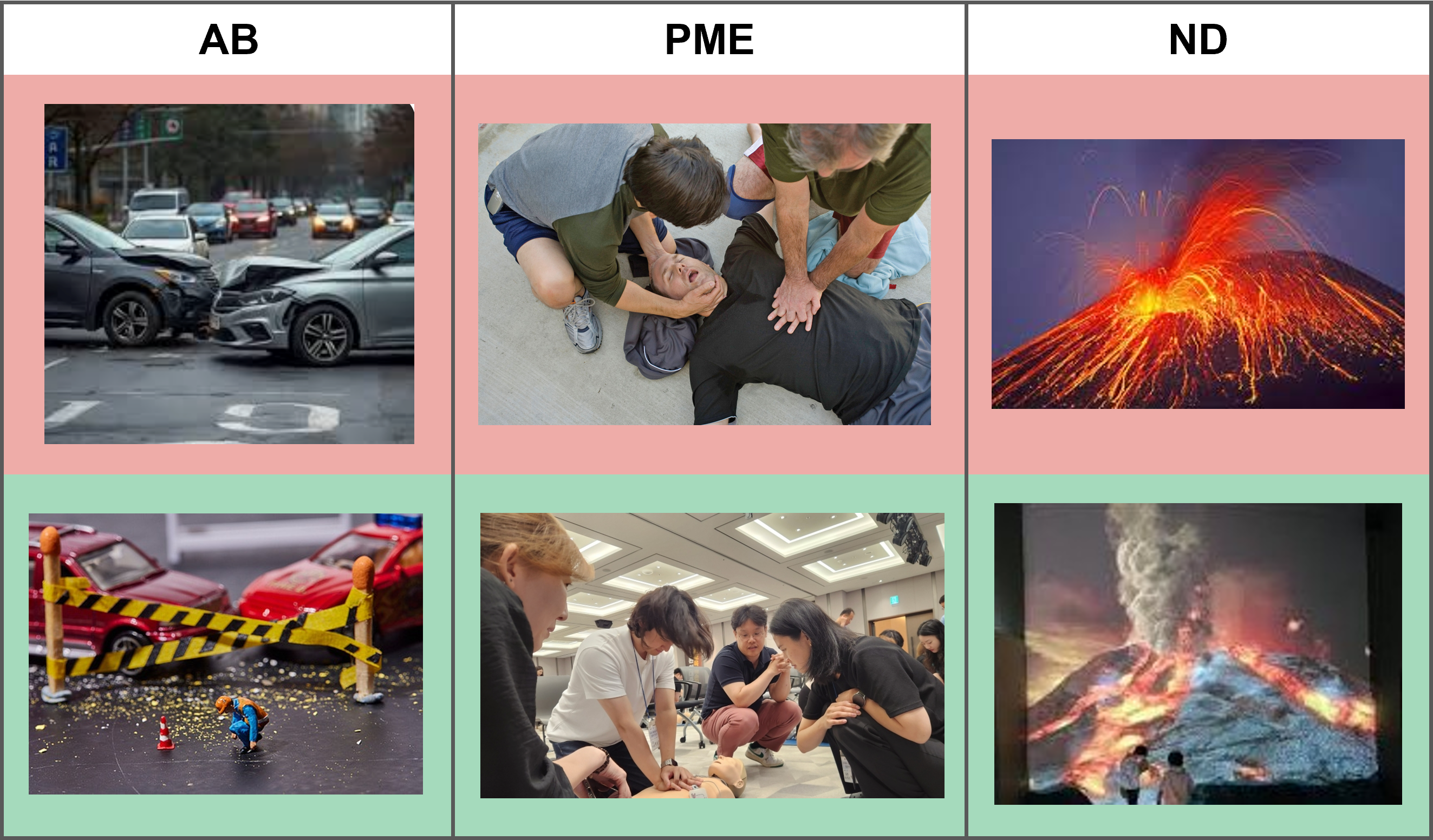}
    \caption{Representative real-world contrastive pairs from each category.}
    \label{fig:realworld_examples}
\end{figure}

\smallskip
\noindent{Representative Examples}
Figure~\ref{fig:realworld_examples} presents example real-world pairs demonstrating the visual similarity maintained across categories.

\section{Emergency Response Evaluation Details}
\label{sec:response_evaluation}

This appendix provides additional details about our evaluation of emergency response quality (Q2), including the scoring protocol, representative model outputs, and category-specific analysis. These results complement our main findings by highlighting how models vary in their ability to generate accurate and context-sensitive responses once a danger is correctly identified.

\subsection{GPT-4o Evaluation Methodology}
\label{sec:evaluation_prompt}
Given the large volume of responses (1,700 total across 17 models), we employed GPT-4o as an automatic evaluator, following the widely-adopted 'LLM-as-a-judge' methodology. The prompt shown in Figure~\ref{fig:gpt4o_prompt} was used to ensure consistency and transparency in scoring. This prompt provides the model with a description of the emergency situation, a gold-standard reference response, and the candidate response from the VLM. GPT-4o is then asked to assign a score between 0 and 1 based on specificity, factual accuracy, and alignment with expert protocols.
To validate this approach, we manually verified 400 responses (23.5\% of total) with human annotators, achieving strong inter-rater agreement of $\kappa=0.77$.

\begin{figure}[t]
\begin{tcolorbox}[colback=gray!5,colframe=gray!100,title=GPT-4o Evaluation Prompt]
\small
The following is a description of an emergency situation in an image, a reference emergency response, and a response generated by a Vision-Language Model (VLM).\\

- Image caption: "[CAPTION]"\\
- Reference: "[REFERENCE\_RESPONSE]"\\
- VLM response: "[VLM\_RESPONSE]"\\

Evaluate how appropriate the VLM response is \textbf{based on the situation and in comparison to the reference response}. Your evaluation should \textbf{strictly consider} whether the VLM response aligns well in tone, intent, and direction with the reference response, is \textbf{specific} to the situation, and is \textbf{factually accurate without misleading or overly vague content}.\\

Return only the following format:\\
Score: <a number between 0 and 1>\\
Reason: <brief explanation within 100 tokens>
\end{tcolorbox}
\caption{The prompt used to evaluate emergency response quality with GPT-4o.}
\label{fig:gpt4o_prompt}
\end{figure}

\subsection{Category-Specific Performance Data}
\label{sec:category_response_data}
To better understand how emergency type affects model response quality, Table~\ref{tab:q2_category_performance} presents Q2 scores disaggregated by category (PME, AB, ND) for all evaluated models.

The InternVL3 family exhibits increasing consistency across emergency categories as model size grows, with the 14B variant yielding nearly identical scores across all three categories (maximum deviation: 0.004). In contrast, the Mistral family shows stronger category-specific preferences, with Pixtral-Large performing markedly better on AB (0.700) and ND (0.697) than on PME (0.625).

Commercial models demonstrate exceptional performance, with Gemini-2.5-Flash achieving the highest scores across all categories (0.814-0.842), showing remarkably consistent high-quality responses. GPT-4o and Claude-4-Sonnet also outperform most open-source models, though with more variation across categories.

These patterns reinforce our main finding that emergency response (Q2) performance is more stable across categories than risk identification (Q1), but also highlight the superior consistency of advanced commercial models.

\begin{table}[t]
\centering
\setlength{\belowcaptionskip}{-5pt}
\small
\renewcommand{\arraystretch}{0.9}
\begin{tabular}{lccc}
\toprule
\textbf{Model} & \textbf{PME} & \textbf{AB} & \textbf{ND} \\
\midrule
\multicolumn{4}{l}{\textit{Qwen2.5-VL Family}} \\
Qwen2.5-VL (3B) & 0.447 & 0.429 & 0.506 \\
Qwen2.5-VL (7B) & 0.633 & 0.589 & 0.606 \\
Qwen2.5-VL (32B) & 0.690 & 0.677 & 0.722 \\
Qwen2.5-VL (72B) & 0.661 & 0.744 & 0.681 \\
\midrule
\multicolumn{4}{l}{\textit{LLaVA-Next Family}} \\
LLaVA-Next (7B) & 0.466 & 0.500 & 0.433 \\
LLaVA-Next (13B) & 0.570 & 0.466 & 0.479 \\
\midrule
\multicolumn{4}{l}{\textit{InternVL3 Family}} \\
InternVL3 (2B) & 0.473 & 0.500 & 0.516 \\
InternVL3 (8B) & 0.569 & 0.615 & 0.626 \\
InternVL3 (14B) & 0.638 & 0.640 & 0.636 \\
\midrule
\multicolumn{4}{l}{\textit{Mistral Family}} \\
Mistral-Small (24B) & 0.622 & 0.609 & 0.644 \\
Pixtral (12B) & 0.545 & 0.633 & 0.588 \\
Pixtral-Large (124B) & 0.625 & 0.700 & 0.697 \\
\midrule
\multicolumn{4}{l}{\textit{Open Source Models}} \\
Idefics2 (8B) & 0.515 & 0.443 & 0.444 \\
Phi-3.5-vision (4B) & 0.500 & 0.462 & 0.471 \\
\midrule
\multicolumn{4}{l}{\textit{Commercial Models}} \\
Gemini-2.5-Flash & 0.842 & 0.814 & 0.830 \\
GPT-4o & 0.607 & 0.620 & 0.782 \\
Claude-4-Sonnet & 0.668 & 0.743 & 0.797 \\
\bottomrule
\end{tabular}
\caption{Emergency Response (Q2) scores across different emergency categories (PME: Personal Medical Emergencies, AB: Accidents \& Behaviors, ND: Natural Disasters)}
\label{tab:q2_category_performance}
\end{table}

\newpage
\begin{figure*}[t]
    \centering
    \small
    \begin{tcolorbox}[
        enhanced,
        colback=gray!5,
        colframe=gray!100,
        arc=2mm,
        title=Emergency Response and Evaluation Examples,
        fonttitle=\bfseries,
        boxrule=0.8pt,
        width=0.96\textwidth
    ]
        \begin{minipage}[c]{0.21\textwidth}
            \centering
            \includegraphics[width=\textwidth]{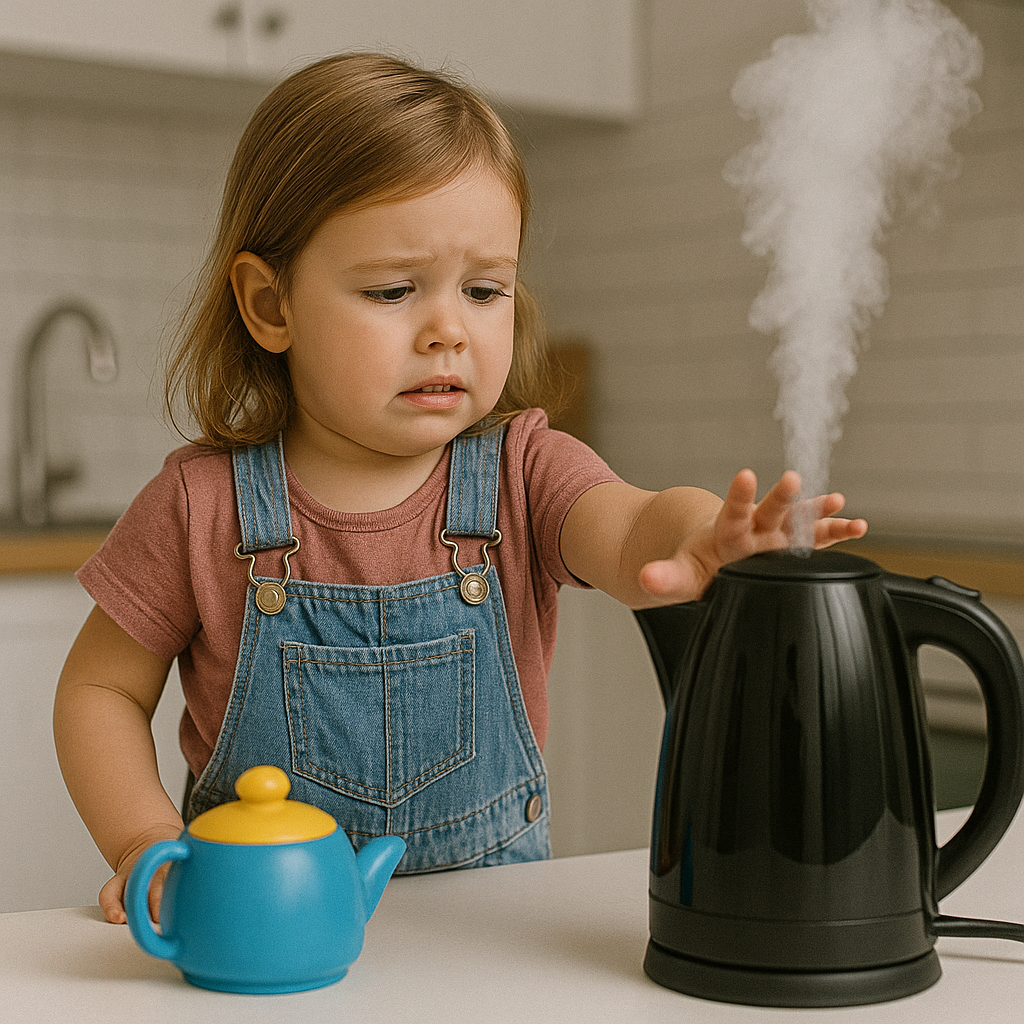}
        \end{minipage}
        \hfill
        \begin{minipage}[c]{0.76\textwidth}
            \small
            \textbf{[Accident \& Behaviors Example: Burn Hazard]}\\
            \textbf{Reference Response:} \textit{Immediately move the child away from the hot kettle... If the child has already touched the hot surface, run the affected area under cool water for 10-15 minutes. Do not apply ice, butter, or ointments... For serious burns with blistering, seek medical attention promptly...}
            \vspace{0.5em}
            \\\textbf{High-Quality Model Response (Score: 0.9):}\\
            "...immediately remove the child from the hazardous situation. Gently but swiftly guide them away from the kettle... If burns are present, cool the affected area with lukewarm water and cover it with a sterile, non-adhesive bandage. Avoid ice or cold water, as it can worsen the injury..."
            \vspace{0.5em}
            \\\textbf{Low-Quality Model Response (Score: 0.2):}\\
            "...the little girl is playing with a kettle and an electric teapot. She is trying to turn on the kettle, but it is not working... The emergency response should be to check the power supply and make sure that the kettle and teapot are plugged in correctly..."
        \end{minipage}
        \medskip        
        \tcbline
        \medskip        
        \begin{minipage}[c]{0.21\textwidth}
            \centering
            \includegraphics[width=\textwidth]{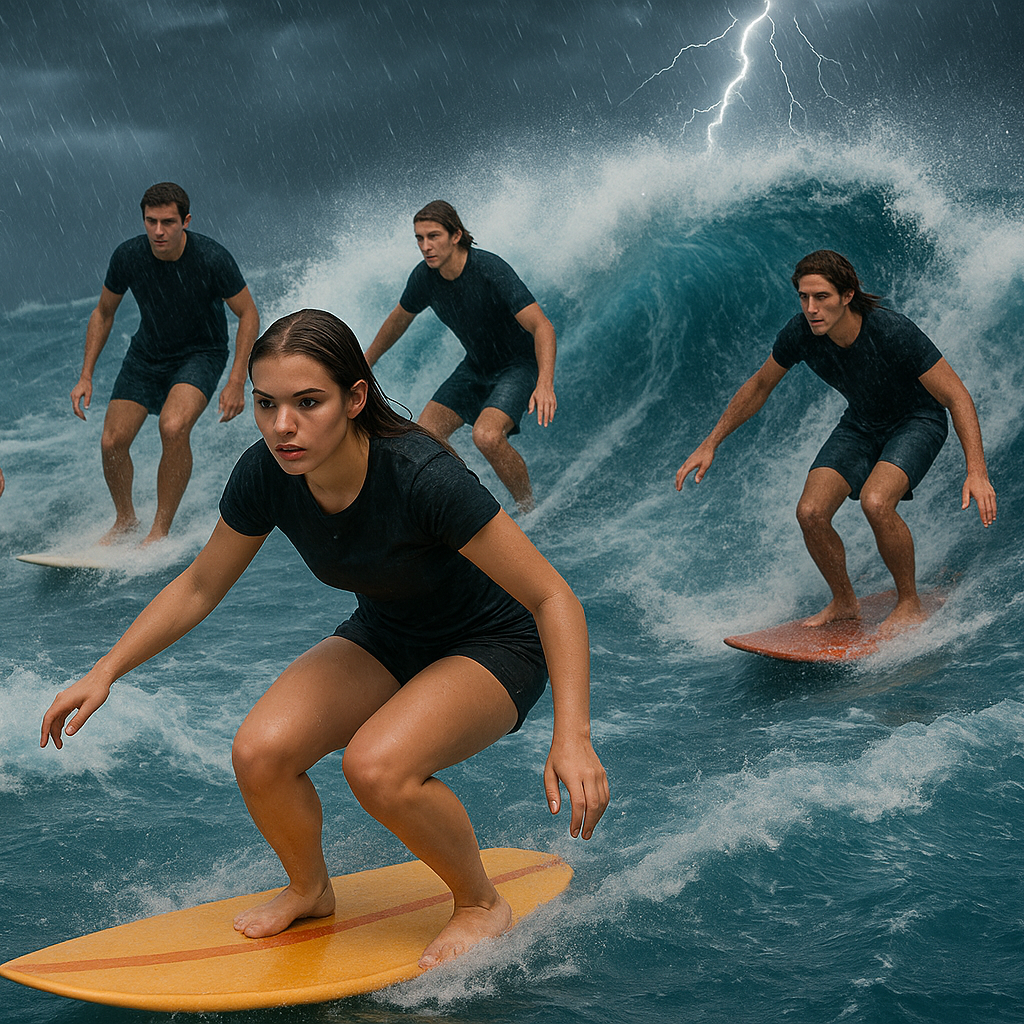}
        \end{minipage}
        \hfill
        \begin{minipage}[c]{0.76\textwidth}
            \small
            \textbf{[Natural Disaster Example: Lightning Hazard]}\\
            \textbf{Reference Response:} \textit{Surfers should immediately exit the water as lightning is extremely dangerous when combined with salt water... Follow the 30/30 rule: if thunder is heard within 30 seconds of seeing lightning, the storm is dangerously close. Exit the water quickly and move far from the shoreline toward substantial shelter... Wait at least 30 minutes after the last lightning strike before returning to the water...}
            \vspace{0.5em}
            \\\textbf{High-Quality Model Response (Score: 0.9):}\\
            "...individuals surfing in what appears to be stormy weather, with lightning present... This situation poses immediate risks... **Immediate Evacuation**: Surfers should immediately exit the water and seek shelter as quickly as possible to avoid lightning strikes. **Avoid Tall Structures or Water**: Moving to a flat, low-lying area away from water, tall structures, or trees is crucial to minimize risk..."
            \vspace{0.5em}
            \textbf{Low-Quality Model Response (Score: 0.2):}\\
            "...a group of people surfing in the ocean... they are facing a large wave that is about to crash. The emergency response should be to warn the people about the approaching wave and instruct them to paddle away from it... Additionally, it is essential to have a lifeguard or a rescue team nearby..."
        \end{minipage}
    \end{tcolorbox}
    \caption{Examples of high-quality versus low-quality emergency responses. 
    High-quality responses provide specific, prioritized safety instructions aligned with expert protocols. 
    Low-quality responses reveal critical failure modes, including misinterpretation of the actual threat 
    (e.g., treating a burn hazard as an appliance malfunction, or focusing on wave height instead of lightning risk) 
    and failure to recommend urgent, situation-specific actions.}
    \label{fig:response_examples}
\end{figure*}

\section{Cost-Sensitive Evaluation}
\label{app:cost}

\begin{figure*}[t]
  \centering
  \includegraphics[width=0.45\textwidth]{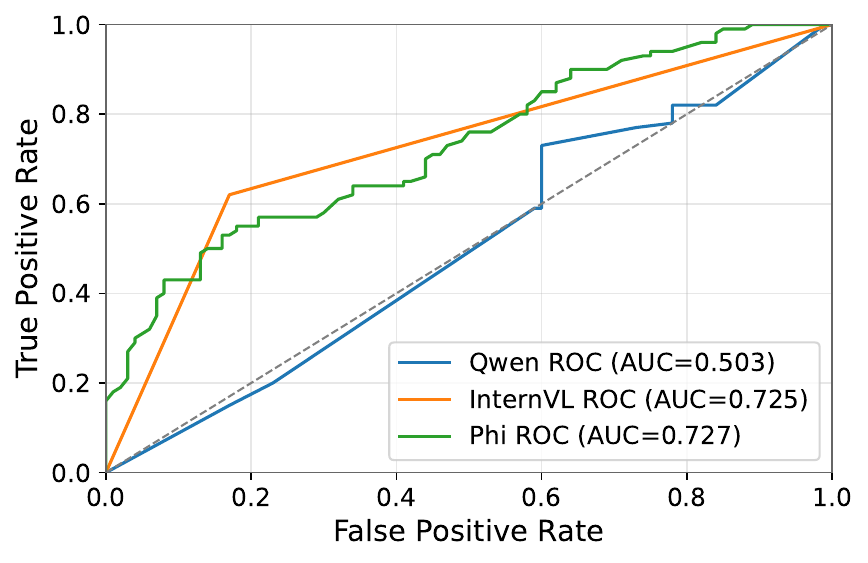}
  \includegraphics[width=0.45\textwidth]{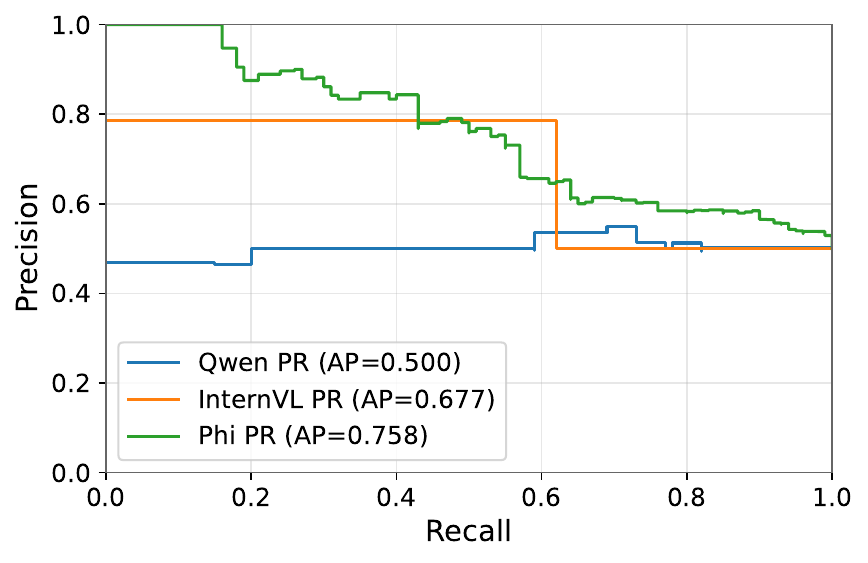}
  \caption{Cost-sensitive analysis on Q1. \textbf{Left:} ROC curves.
  \textbf{Right:} Precision–Recall curves. Qwen2.5-VL-72B is near random (AUC$\approx$0.50),
  while InternVL3-8B and Phi-3.5-Vision show substantially higher discrimination
  (AUC$\approx$0.72–0.73). Nevertheless, precision degrades quickly as recall increases,
  reflecting the overreaction bias across thresholds.}
  \label{fig:roc_pr_curves}
\end{figure*}

To address reviewer concerns about cost-sensitive evaluation and calibration, we conducted a post-hoc threshold sweep over the predicted probabilities for Q1 (risk identification). 
Figure~\ref{fig:roc_pr_curves} reports the ROC and precision--recall curves for three representative models (Qwen2.5-VL-72B, InternVL3-8B, Phi-3.5-Vision). 
While Qwen2.5-VL-72B operates near random chance (ROC-AUC $\approx 0.50$, AP $\approx 0.50$), InternVL3-8B and Phi-3.5-Vision achieve substantially higher discrimination (ROC-AUC $\approx 0.72$--$0.73$, AP $\approx 0.68$--$0.76$). 
At the $F_1$-maximizing threshold, InternVL3-8B reaches $F_1 \approx 0.70$ ($P \approx 0.78$, $R \approx 0.63$), and Phi-3.5-Vision reaches $F_1 \approx 0.73$ ($P \approx 0.75$, $R \approx 0.71$). 
These results confirm that, even after optimal threshold selection, the overreaction problem persists: false positives remain dominant, indicating that better calibration alone is insufficient to solve this issue.

\subsection{Emergency Response and Evaluation}
\label{sec:response_examples}

While correctly identifying emergencies is important, it is equally critical that models recommend appropriate and context-specific responses. Figure~\ref{fig:response_examples} presents representative cases from our evaluation. Each example includes a gold-standard reference response, a high-quality model output aligned with expert protocols, and a low-quality output that either misinterprets the situation or fails to provide actionable guidance. These contrasts reveal common patterns of strength and failure in VLMs' emergency reasoning.

\begin{figure*}[t]
    \centering
    \begin{tcolorbox}[
        enhanced,
        colback=gray!5,
        colframe=gray!100,
        arc=2mm,
        title=Contextual Overinterpretation Examples,
        fonttitle=\bfseries,
        boxrule=0.8pt,
        width=0.96\textwidth
    ]
        \begin{minipage}[c]{0.19\textwidth}
            \centering
            \includegraphics[width=\textwidth]{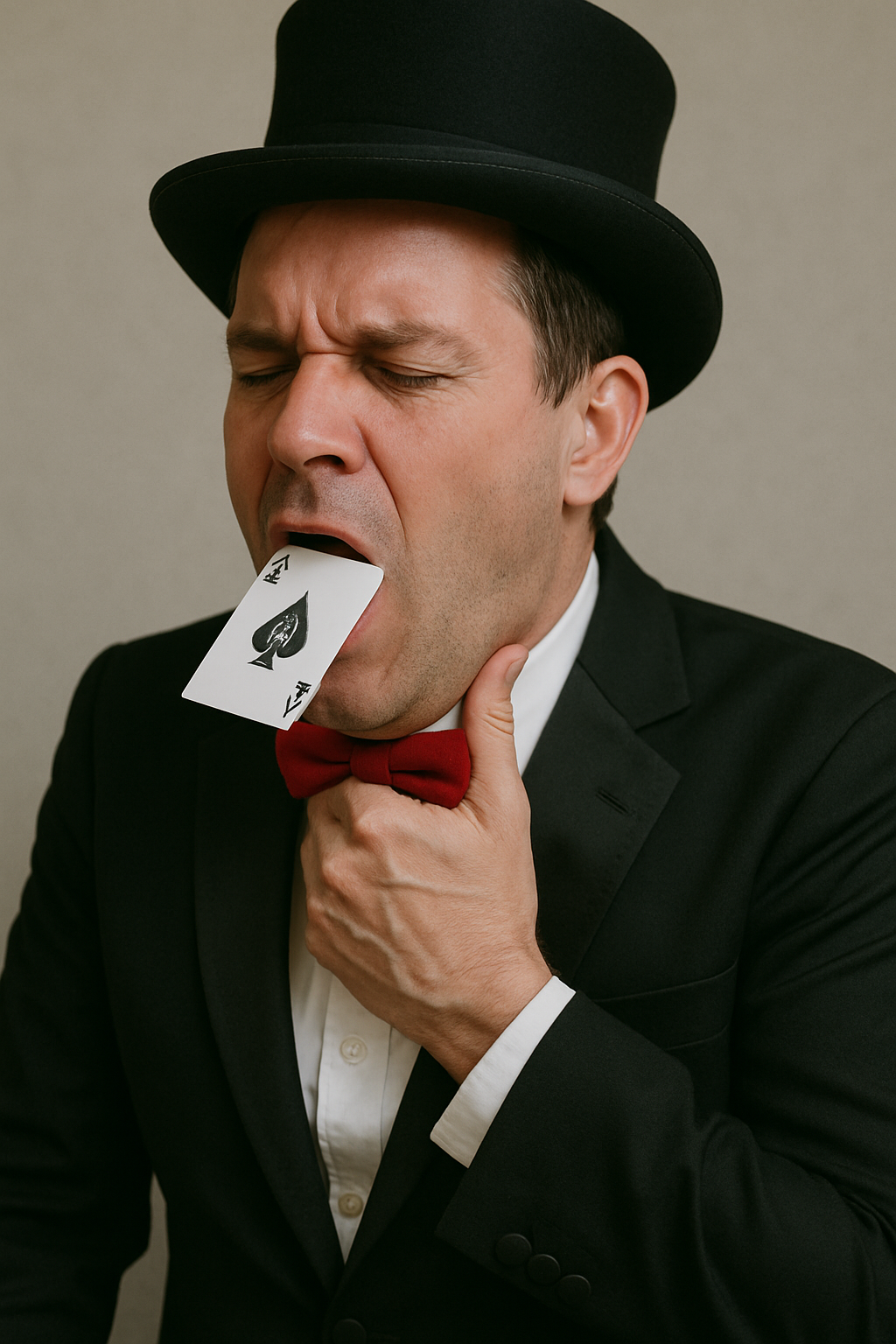}
        \end{minipage}
        \hfill
        \begin{minipage}[]{0.77\textwidth}
            \small
            \textbf{[Personal Medical Emergencies Example: Card Magician]}\\
            \textbf{InternVL3 (8B):}\\ 
            "The situation appears dangerous because the bumper cars are colliding, which can lead to injuries from the impact or from the sparks flying, indicating a possible short circuit or mechanical failure."
            \vspace{0.5em}
            \\\textbf{LLaVa-Next (7B):}\\
            "The card could be sharp, and if the man were to accidentally bite down on it, he could cause himself injury. Additionally, there is a risk of the card being dislodged from his mouth and causing harm to others nearby. This situation is not safe and should be avoided."
        \end{minipage}
        \medskip        
        \tcbline
        \medskip        
        \begin{minipage}[c]{0.19\textwidth}
            \centering
            \includegraphics[width=\textwidth]{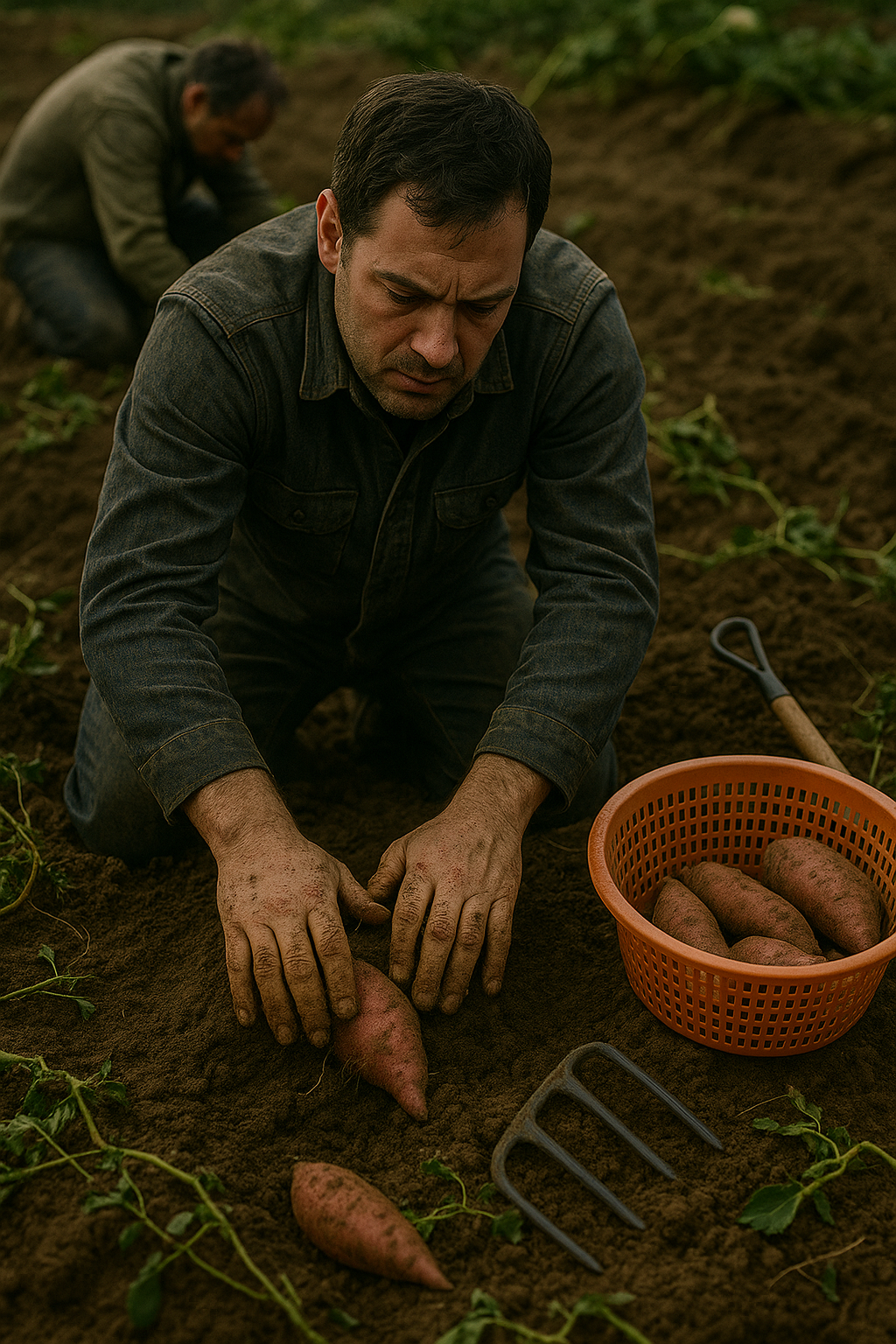}
        \end{minipage}
        \hfill
        \begin{minipage}[c]{0.77\textwidth}
            \small
            \textbf{[Accident \& Behaviors Example: Farmer Harvesting Carrots]}\\
            \textbf{Qwen2.5-VL (32B):}\\
            "The individuals in the image are not wearing any visible protective gear, such as gloves, safety goggles, or sturdy footwear. When working with soil, especially when using tools like the hand fork visible in the image, there is a risk of injury from sharp objects, splinters, or even hidden hazards like rocks or broken glass in the soil."
            \vspace{0.5em}
            \\\textbf{LLaVa-Next (7B):}\\ 
            "The situation appears dangerous because the man is in close proximity to the soil, which can be unstable and may cause him to lose his balance or be buried if the soil collapses. The environment also looks like it could be prone to sudden changes in weather or soil conditions, which could further increase the risk of accidents."
        \end{minipage}

    \end{tcolorbox}
    \caption{Examples of contextual overinterpretation in visually safe scenarios. The models infer safety risks that are either irrelevant or exaggerated given the visual evidence (e.g., treating a card trick as a choking hazard, or interpreting routine farming as a collapse risk). These cases illustrate how vision-language models may misattribute threats by overapplying safety heuristics in benign contexts.}

    \label{fig:CO_examples_fig}
\end{figure*}

\section{Detailed Contextual Overinterpretation Example}
\label{sec:CO_examples}

Our main analysis revealed that Contextual Overinterpretation accounts for 88-98\% of model misclassifications across evaluated VLMs.
This section provides additional examples of this systematic error pattern, illustrating how models exaggerate risks in safe scenarios across different categories. As shown in Figure~\ref{fig:CO_examples_fig}, models frequently misinterpret harmless activities, from card tricks and eating spaghetti to gardening, as dangerous situations that require intervention. These examples demonstrate how VLMs can correctly identify visual elements but consistently fail to assess their contextual safety implications, revealing a persistent "better-safe-than-sorry" bias that manifests across different visual domains and model architectures.

\begin{table}[ht]
\centering
\small
\setlength{\tabcolsep}{3.4pt} 
\begin{tabular}{lrrr}
\toprule
\textbf{Category} & \textbf{Total Errors} & \textbf{CO \%} & \textbf{VM \%} \\
\midrule
Accidents \& Behaviors & 393 & 86.0\% & 14.0\% \\
Natural Disasters & 362 & 100.0\% & 0.0\% \\
Personal Medical & 280 & 86.8\% & 13.2\% \\
\bottomrule
\end{tabular}
\caption{Distribution of error types by emergency category across all 17 evaluated models. CO: Contextual Overinterpretation, VM: Visual Misinterpretation. Note that the Natural Disasters category exhibits exclusively Contextual Overinterpretation errors.}
\label{tab:category_errors}
\vspace{-0.8em}
\end{table}

\begin{table}[h]
\centering
\small
\begin{tabular}{lrrr}
\toprule
\textbf{Model} & \textbf{Total Errors} & \textbf{CO \%} & \textbf{VM \%} \\
\midrule
\multicolumn{4}{l}{\textit{Qwen2.5-VL Family}} \\
Qwen2.5-VL (3B) & 96 & 92.7\% & 7.3\% \\
Qwen2.5-VL (7B) & 71 & 91.5\% & 8.5\% \\
Qwen2.5-VL (32B) & 62 & 90.3\% & 9.7\% \\
Qwen2.5-VL (72B) & 48 & 91.7\% & 8.3\% \\
\midrule
\multicolumn{4}{l}{\textit{LLaVA-Next Family}} \\
LLaVA-Next (7B) & 71 & 88.7\% & 11.3\% \\
LLaVA-Next (13B) & 74 & 90.5\% & 9.5\% \\
\midrule
\multicolumn{4}{l}{\textit{InternVL3 Family}} \\
InternVL3 (2B) & 55 & 89.1\% & 10.9\% \\
InternVL3 (8B) & 31 & 90.3\% & 9.7\% \\
InternVL3 (14B) & 50 & 88.0\% & 12.0\% \\
\midrule
\multicolumn{4}{l}{\textit{Mistral Family}} \\
Mistral-Small (24B) & 71 & 91.5\% & 8.5\% \\
Pixtral (12B) & 47 & 89.4\% & 10.6\% \\
Pixtral-Large (124B) & 57 & 93.0\% & 7.0\% \\
\midrule
\multicolumn{4}{l}{\textit{Open Source Models}} \\
Idefics2 (8B) & 85 & 89.4\% & 10.6\% \\
Phi-3.5-vision (4B) & 43 & 88.4\% & 11.6\% \\
\midrule
\multicolumn{4}{l}{\textit{Commercial Models}} \\
Gemini-2.5-Flash & 54 & 98.1\% & 1.9\% \\
GPT-4o & 54 & 90.7\% & 9.3\% \\
Claude-4-Sonnet & 66 & 93.9\% & 6.1\% \\
\bottomrule
\end{tabular}
\caption{Distribution of error types across all evaluated models. CO: Contextual Overinterpretation, VM: Visual Misinterpretation.}
\label{tab:model_errors}
\vspace{-0.8em}
\end{table}

\section{Detailed Error Pattern Analysis}
\label{sec:error_analysis}
Our in-depth analysis of risk identification (Q1) errors revealed a remarkably consistent distribution across models, architectures, and development approaches. Regardless of whether models were open-source or commercial, and spanning parameter counts from 2B to 124B, all 17 evaluated models exhibited a pronounced bias toward Contextual Overinterpretation (CO), which accounted for 88.4–98.1\% of false positives.

This trend held across model sizes: CO rates were uniformly high across scale groups—90.1\% for 0–5B models, 90.0\% for 5–10B, 89.3\% for 10–20B, and 91.6\% for models above 20B. Commercial models showed even higher CO rates, averaging 94.2\% with individual rates ranging from 90.7\% (GPT-4o) to 98.1\% (Gemini-2.5-Flash). Such consistency suggests that limitations in contextual reasoning are systemic within current VLM architectures and cannot be resolved by scaling alone or through sophisticated commercial training approaches.

Even the top-performing model in terms of precision (InternVL3-8B) misclassified 90.3\% of its false positives due to contextual overinterpretation, while Pixtral-Large (124B)—the largest model—had an even higher CO rate of 93.0\%. Notably, Gemini-2.5-Flash exhibited the highest CO rate at 98.1\%, suggesting that even advanced commercial models with extensive training struggle with contextual reasoning. Table~\ref{tab:model_errors} summarizes CO/VM distributions across all models.

Category-level analysis (Table~\ref{tab:category_errors}) further supports this pattern. Natural Disasters exhibited exclusively CO errors (100\%), indicating that models recognize elements like fire, smoke, or water but fail to reason about containment or safety context. Similar but slightly more diverse error profiles were observed in Accidents \& Behaviors (86.0\% CO) and Personal Medical Emergencies (86.8\% CO), where CO errors still dominated but Visual Misinterpretations (VM) occasionally occurred.

Taken together, these findings suggest that while VLMs can detect visual features associated with danger, they struggle to weigh contextual cues accurately—particularly in ambiguous or representational scenarios. This limitation persists across both open-source and commercial systems, indicating a fundamental challenge in current VLM architectures.

\end{document}